\documentclass{ifacconf}
\usepackage{color}
\usepackage{url}
\newtheorem{theorem}{Theorem}[section]
\usepackage{amsmath} 
\usepackage{amssymb}
\usepackage{graphicx}      
\usepackage{natbib}        

\allowdisplaybreaks

\begin{document}
\begin{frontmatter}

\title{Closed-Loop Model Identification and MPC-based Navigation of Quadcopters: A Case Study of Parrot Bebop 2}

\author{Mohsen Amiri and Mehdi Hosseinzadeh}

\address{School of Mechanical and Materials Engineering, Washington State University, Pullman, WA 99164, USA \\
E-mail: mohsen.amiri@wsu.edu;~mehdi.hosseinzadeh@wsu.edi.}

\begin{abstract}  
The growing potential of quadcopters in various domains, such as aerial photography, search and rescue, and infrastructure inspection, underscores the need for real-time control under strict safety and operational constraints. This challenge is compounded by the inherent nonlinear dynamics of quadcopters and the on-board computational limitations they face. This paper aims at addressing these challenges. First, this paper presents a comprehensive procedure for deriving a linear yet efficient model to describe the dynamics of quadrotors, thereby reducing complexity without compromising efficiency. Then, this paper develops a steady-state-aware Model Predictive Control (MPC) to effectively navigate quadcopters, while guaranteeing constraint satisfaction at all times. The main advantage of the steady-state-aware MPC is its low computational complexity, which makes it an appropriate choice for systems with limited computing capacity, like quadcopters. This paper considers the Parrot Bebop 2 as the running example, and experimentally validates and evaluates the proposed algorithms. 
\end{abstract}


\end{frontmatter}

\section{Introduction}

\noindent\textbf{Motivation:} Quadcopters have demonstrated proficiency in hovering and have found diverse applications across fields such as aerial photography, surveillance, precision agriculture, and package delivery \citep{Gandhi2018,HosseinzadehMED2018,Duggal2016,Kumar2018}. Nevertheless, their nonlinear dynamics pose significant challenges for modeling and control, and specifically for addressing safety and operational constraints \citep{shauqee2021quadrotor,bouabdallah2004pid,idrissi2022review}. In particular, the inherent nonlinearity of quadcopter dynamics, arising from intricate aerodynamic interactions and dynamic coupling between motion axes, complicates the accurate identification of models and the development of safe and reliable control strategies.

A further significant challenge in navigating quadcopters, in practical, in terms of precise position tracking, is that they are often designed to keep weight and energy consumption as low as possible \citep{Ichnowski2019,Hosseinzadeh2022_ROTEC,hosseinzadeh2023TAC}; thus, the computing power of quadcopters is naturally limited, while multiple computationally expensive diagnostics and control functions must be executed at any time instant. This underlines the importance of identifying a simple, yet accurate, model to describe the dynamics of quadcopters, and of developing a control scheme that is capable of guaranteeing constraint satisfaction despite limitations of computing power.

\noindent\textbf{Prior Work:} Several previous studies have attempted to model quadcopter dynamics, considering aerodynamic effects, rotor dynamics, and environmental disturbances; see e.g., \citep{Wang2016,Khodja2017,Zhao2014,Saif2022,tang2017nonlinear}. Identification methods based on artificial intelligence have been considered in \citep{kantue2018nonlinear,pairan2020neural,wang2023nonlinear}. The above-mentioned efforts provide a nonlinear model to describe the dynamical behavior of quadcopters. One inevitable issue with the nonlinear models is that controllers designed based on them tend to be complex and computationally intensive \citep{wang2017nonlinear}, challenging their practical implementation, in particular, when the available computing power is limited.  To overcome these limitations, specifically for navigation purposes, a shift towards simpler nonlinear models (e.g., \citep{pinto2020high}) and linear models (e.g., \citep{HosseinzadehMED2018,Santos2017_2,Santos2017,Santana2014,Santos2019} has been made in recent years. A linear model offers a more tractable framework for modeling and control, facilitating the development of precise navigation algorithms while addressing safety constraints. To the best of our knowledge, prior work on identifying a linear model to describe the dynamical behavior of quadcopters lacks a detailed procedure and/or fails to effectively account for output constraints; in particular, most of the existing articles perform an open-loop identification that is not appropriate for indoor identification.




Once a dynamical model is identified, one possible approach \citep{Huang2000,Huang2003} to evaluate its reliability is to investigate the performance of a Model Predictive Control (MPC) scheme developed based on that model. Ensuring constraint satisfaction is crucial for safe and efficient operation of quadcopters \citep{dubay2018distributed}, and MPC has shown promise in this domain by enabling online decision-making based on predictive models \citep{camacho2007model}. However, the computational challenges associated with MPC implementation pose significant barriers. Several approaches have been proposed to mitigate this challenge, including explicit MPC \citep{alessio2009survey,kvasnica2019complexity} and self- and event-triggered MPC \citep{sun2019robust,wang2020event,feller2018sparsity}. Anytime MPC \citep{feller2018sparsity} ensures stability with minimal iterations, although it cannot guarantee constraint satisfaction at all times. Converting MPC problem into the evolution of a continuous-time system has been introduced in \citep{nicotra2018embedding,Hosseinzadeh2023RobustTermination}, while its discrete-time implementation is still challenging.

Shortening the prediction horizon, as the most intuitive approach to reduce the computational cost of MPC, has been considered in \citep{sawma2018effect,Pannek2011}. To address feasibility issues with small prediction horizons, \citep{Ferramosca2009,Limon2008,Ferramosca2008} introduced modifications to the conventional MPC; note that to ensure output tracking, such modifications need prior knowledge about the steady state and steady input corresponding to any given reference, which is computationally demanding \citep{Yousfi1991,Muske1993,Shead2010} and practically unrealistic, in particular, when the reference is unknown \textit{a priori}.

Despite the above-mentioned efforts, achieving computational efficiency, while addressing performance objectives and enforcing constraint satisfaction remains a key research focus. Recently, the authors have proposed a steady-state-aware MPC framework in \citep{SSMPC} which is appropriate for systems with limited computing power. The main feature of the steady-state-aware MPC is that, given a desired steady state and input, it guarantees that the state and input of the system converge to the desired steady values, while ensuring output tracking and constraint satisfaction at all times, without requiring any prior knowledge about the reference. Simulation studies demonstrated the capability of the steady-state-aware MPC in guaranteeing output tracking and steady-state convergence, as well as enforcing constraint satisfaction at all times, all under the limitations on available computing power. However, experimental validations are still required to establish the steady-state-aware MPC as a standard approach for safe and reliable control of safety-critical systems with limited computational resources.



\begin{figure}
\begin{center}
\includegraphics[width=5.4cm]{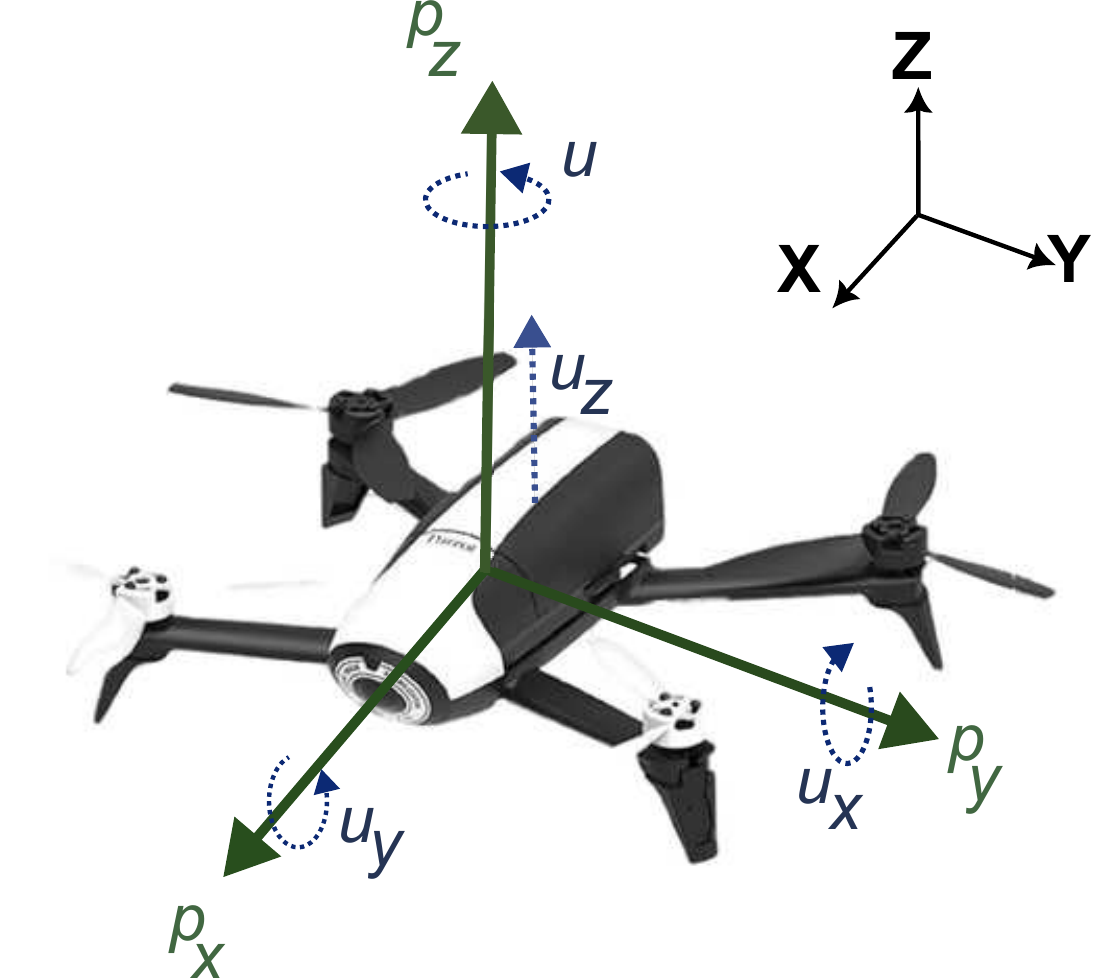}    
\caption{Parrot Bebop 2 and considered reference and body frames; note that to make the measurements more relevant, the considered body frame is different than the one in the ``Parrot Drone Support from MATLAB" package \citep{MATLAB}.} 
\label{fig:Parrot}
\end{center}
\end{figure}

\noindent\textbf{Goal:} In a nutshell, the main goal of this paper is to systematically address the intricacies of quadcopter dynamics and computational limitations. First, we advocate for a physically informed modeling approach, and identify a simple linear yet efficient model for a quadcopter to enable efficient onboard computation and safe control. Then, we develop a steady-state-aware MPC to safely navigate the quadcopter, while managing constraints effectively despite limited available computational resources. We consider a Parrot Bebop 2 (see Fig. \ref{fig:Parrot}) as a running example throughout the paper, and experimentally evaluate and validate the proposed methodology.


\noindent\textbf{Organization:} The rest of this paper is organized as follows. The presentation of the model structure followed by identification procedure is discussed in Section \ref{sec:Modeling}. Section \ref{sec:Control} develops a steady-state-aware MPC and experimentally assesses its effectiveness on Parrot Bebop2 drone. Finally, Section \ref{sec:Conclusion} concludes this paper.

\noindent\textbf{Notation:} We denote the set of real numbers by $\mathbb{R}$, the set of positive real numbers by $\mathbb{R}_{>0}$, and the set of non-negative real numbers by $\mathbb{R}_{\geq0}$. Similarly, we use $\mathbb{Z}$ to denote the set of integer numbers, $\mathbb{Z}_{>0}$ to denote the set of positive integer numbers, and $\mathbb{Z}_{\geq0}$ to denote the set of non-negative integer numbers. We denote the transpose of matrix $A$ by $A^\top$. Also, $A\succ0$ ($A\succeq0$) indicates that $A$ is positive definite (positive semi-definite). Given $x\in\mathbb{R}^n$ and $Q\succeq0$, we have $\left\Vert x\right\Vert_Q^2=x^\top Qx$. We use $t$ to denote continuous time and $k$ to denote sampling instants. The notation $\text{diag}\{a_1,\cdots,a_n\}$ indicates a $n\times n$ diagonal matrix with elements $a_1,\cdots,a_n$ on the main diagonal. Finally, $\mathbf{0}$ denotes the zero matrix with appropriate dimension.

\section{Closed-Loop Model Identification}\label{sec:Modeling}

\subsection{Model Structure}
Suppose that the yaw angle of the quadcopter is very small; see Subsection \ref{sec:IdentificationProcedure} for details on how one can regulate the yaw angle to zero. Taking inspiration from \citep{pinto2020high,Santos2019}, we propose the following structure to describe the position dynamics of a quadcopter:
\begin{subequations}\label{eq:System1}
\begin{align}
\ddot{p}_x+\alpha_x\dot{p}_x=&\beta_xu_x,\\
\ddot{p}_y+\alpha_y\dot{p}_y=&\beta_yu_y,\\
\ddot{p}_z+\alpha_z\dot{p}_z=&\beta_zu_z,
\end{align}
\end{subequations}
where $p_x,p_y,p_z\in\mathbb{R}$ are  X, Y, and Z positions of the quadcopter in the global Cartesian coordinate system, $\alpha_x,\alpha_y,\alpha_z,\beta_x,\beta_y,\beta_z\in\mathbb{R}$ are system parameters, and $u_x,u_y,u_z\in\mathbb{R}$ are control inputs on X, Y, and Z directions.

We use the ``Parrot Drone Support from MATLAB" \citep{MATLAB} to send control commands to the Parrot Bebop 2 drone. In this framework, $u_x$ is the pitch angle (in [rad]), $u_y$ is the roll angle (in [rad]), and $u_z$ is the vertical velocity (in [m/s]); see Fig. \ref{fig:Parrot}.

The dynamical model \eqref{eq:System1} can be expressed via the following state-space equations:
\begin{subequations}\label{eq:StateEqaution}
\begin{align}
\dot{x}=&\begin{bmatrix}
0 & 1 & 0 & 0 & 0 & 0 \\
0 & -\alpha_x  & 0 & 0 & 0 & 0 \\
0 & 0 & 0 & 1 & 0 & 0 \\
0 & 0 & 0 &-\alpha_y & 0 & 0 \\
0 & 0 & 0 & 0 & 0 & 1 \\
0 & 0 & 0 & 0 & 0 & -\alpha_z
\end{bmatrix} x+\begin{bmatrix}
0 & 0 & 0  \\
\beta_x & 0 & 0  \\
0 & 0 & 0  \\
0 & \beta_y & 0  \\
0 & 0 & 0  \\
0 & 0 & \beta_z 
\end{bmatrix}u,\\
y=&\begin{bmatrix}1&0&0&0&0&0\\0&0&1&0&0&0\\0&0&0&0&1&0\end{bmatrix}x,
\end{align}
\end{subequations}
where $x=\begin{bmatrix}p_x & \dot{p}_x & p_y & \dot{p}_y & p_z & \dot{p}_z \end{bmatrix}^\top$ and $u=\begin{bmatrix}u_x& u_y& u_z \end{bmatrix}^\top$.

Note that model \eqref{eq:StateEqaution} is consistent with practical observations. First, dynamics along X, Y, and Z directions are decoupled, as one can control the position of the quadcopter along X, Y, and Z directions separately. Second, when $\dot{p}_x\neq0$ ($\dot{p}_y\neq0$ and $\dot{p}_z\neq0$) and $u_x=0$ ($u_y=0$, $u_z=0$), the quadcopter continues moving in x direction (y and z directions) while its velocity along that direction decreases to zero; from this observation, we anticipate that $\alpha_x$, $\alpha_y$, and $\alpha_z$ are positive (i.e., $\alpha_x,\alpha_y,\alpha_z\in\mathbb{R}_{>0}$). Third, when pitch and roll angles are small, the quadcopter behaves as a second-order system.

\subsection{Identification Procedure}\label{sec:IdentificationProcedure}
According to model \eqref{eq:System1}, it is possible to obtain the system parameters (i.e., $\alpha_i,\beta_i,~i\in\{x,y,z\}$) by applying a control input signal and measuring its response; this method is called the open-loop identification and is followed widely in prior work, e.g., \citep{pinto2020high,Santana2014,Santos2019}. Although the quadcopters are stable, open-loop identification is not appropriate for indoor identification, as the spatial constraints and limits cannot be enforced. A different approach, which is called the closed-loop identification \citep{Forssell1999,Forssell2000}, is to identify the system parameters when a controller is utilized to control the position and angles of the quadcopter. In addition to enforcing output constraints, closed-loop identification allows us \citep{Hof1998} to limit the operating points of the quadcopter to a region in which the quadcopter presents a linear behavior (i.e., keeping the pitch and roll angles small). 

Note that since the measurement noise in motion capture systems used in indoor applications is very small, the control input due to the measurement noise will be small as well. Thus, one can use the closed-loop identification approach without being worried about the correlation between the measurement noise and the control input, which is discussed in \citep{Forssell1999}.

In what follows, we provide a systematic procedure for closed-loop identification of quadcopters. 


\begin{itemize}
    \item\textbf{Network Establishment:} The experimental setup should consist of a motion capture system that measures the position of the quadcopter, a computing unit that analyzes the camera data and computes the control commands, and a communication channel that sends the control commands to the quadcopter.

Our experimental setup is shown in Fig. \ref{fig:network}. We use \texttt{OptiTrack} system with ten \texttt{Prime$^\text{x}$ 13} cameras operating at a frequency of 120 Hz; these cameras provide a 3D accuracy of $\pm0.02$ millimeters which is acceptable for identification purposes. The computing unit is a $13^{\text{th}}$ Gen $\text{Intel}^{\text{\textregistered}}$ $\text{Core}^{\text{\texttrademark}}$ i9-13900K processor with 64GB RAM, on which the software \texttt{Motive} is installed to analyze and interpret the camera data. We use the ``Parrot Drone Support from MATLAB" package \citep{MATLAB}, and send the control commands to the Parrot Bebop 2 via WiFi and by using the command \texttt{move($\cdot$)}. It should be mentioned that the communication between \texttt{Motive} and MATLAB is done through User Datagram Protocol (UDP) communication using the \texttt{NatNet} service.

\begin{figure}
\begin{center}
\includegraphics[width=\columnwidth]{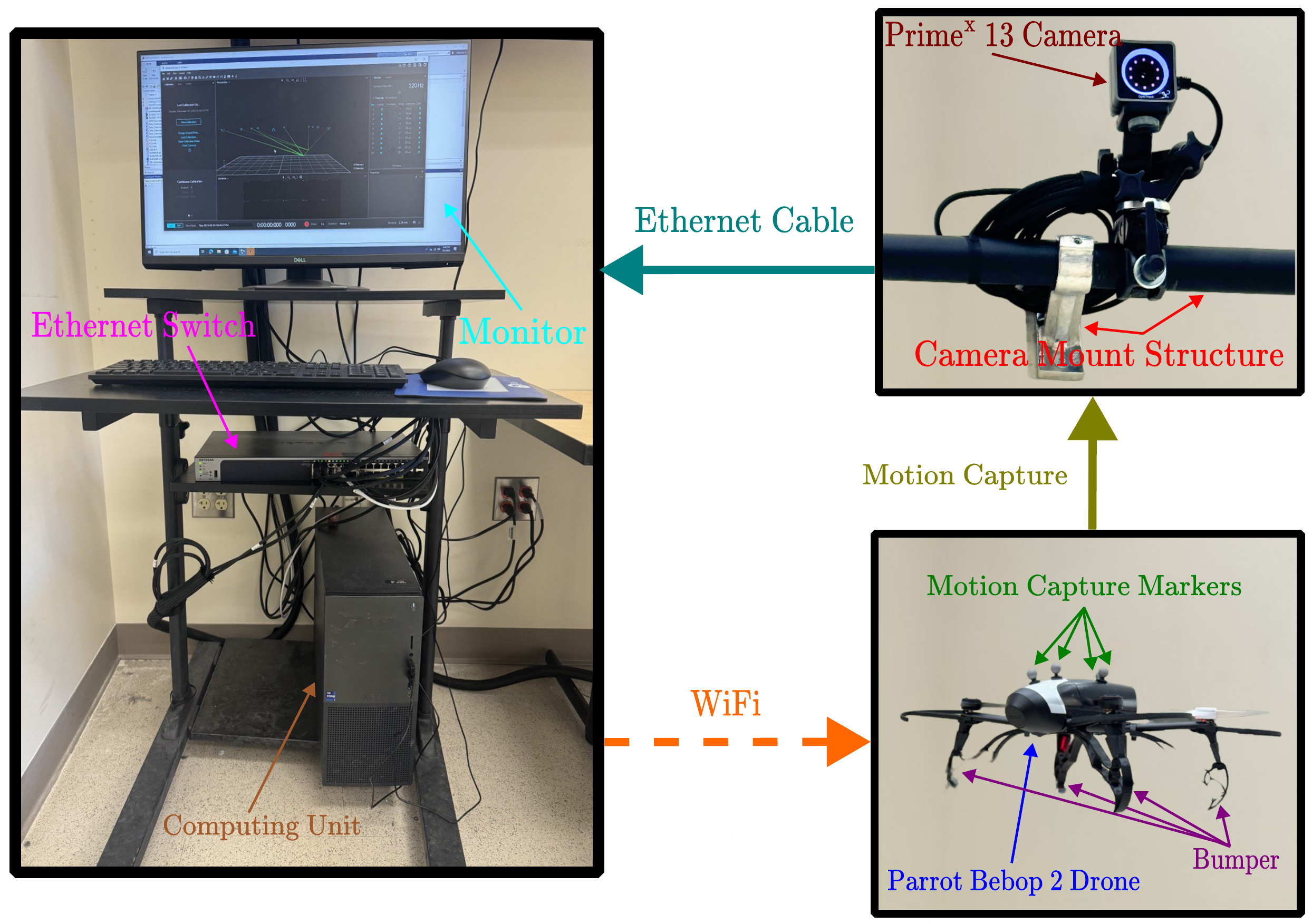}  \\
\caption{Overview of the experimental setup utilized to evaluate and validate the proposed methods.}
\label{fig:network}
\end{center}
\end{figure}

\item \textbf{Nominal Controller:} One can use a  Proportional-Derivative (PD) control law to control the position of the quadcopter and its yaw angle, as follows:
\begin{subequations}\label{eq:PDController}
    \begin{align}
    u_x=&-k_p^x\left(p_x-p_x^d\right)-k_d^x\left(\dot{p}_x-\dot{p}_x^d\right),\\
    u_y=&-k_p^y\left(p_y-p_y^d\right)-k_d^y\left(\dot{p}_y-\dot{p}_y^d\right),\\
    u_z=&-k_p^z\left(p_z-p_z^d\right)-k_d^z\left(\dot{p}_z-\dot{p}_z^d\right),\\
    u_\psi=&-k^\psi_p\left(\psi-\psi^d\right)-k^\psi_d\left(\dot{\psi}-\dot{\psi}^d\right),
    \end{align}
\end{subequations}
where $\psi$ is the yaw angle, $p_x^d,p_y^d,p_z^d\in\mathbb{R}$ are the desired X, Y, and Z trajectories, $\psi^d=0$ is the desired yaw angle, $k^x_p,k_p^y,k^z_p,k^\psi_p\in\mathbb{R}_{>0}$ are proportional gains, and $k^x_d,k_d^y,k^z_d,k^\psi_d\in\mathbb{R}_{>0}$ are derivative gains.

\begin{table}[!t]
\caption{Parameters of the PD controllers used to control the Parrot Bebop 2 drone.}
\centering
\begin{tabular}{c|c}
Parameter &         Identified Value   \\
\hline\hline
$k_p^x$ & 0.050 \\
\hline
$k_p^y$ & 0.050 \\
\hline
$k_p^z$ & 1.700 \\
\hline
$k_p^\psi$ & 1.000 \\
\hline
$k_d^x$ & 0.065 \\
\hline
$k_d^y$ & 0.065 \\
\hline
$k_d^z$ & 0.200 \\
\hline
$k_d^\psi$ & 0.800 \\
\end{tabular}
\label{tab:PDParamteres}
\end{table}

In our experiments, we select the parameters of the PD controllers as in Table \ref{tab:PDParamteres}. Note that this selection ensures that the yaw angle $\psi$ remains around zero while Parrot Bebop 2  moves toward the desired position $(P_x^d,P_y^d,P_z^d)$. Also, note that since the \texttt{OptiTrack} system does not provide the velocities along X, Y, and Z directions, to implement the PD controllers \eqref{eq:PDController}, we use the Newton's difference quotient to compute the velocities along all directions.


\item\textbf{Trajectory Determination For Closed-Loop Identification:} The most intuitive way to identify a system is to apply a combination of sinusoidal inputs at different frequencies and observe the output of the system. It is well-known that if the input to a linear system is sinusoidal, the output will also be sinusoidal, and at exactly the same frequency as the input. This property, which is called sinusoidal fidelity, provides insights on how to determine the desired output trajectory for closed-loop identification and on how to identify the operating region in which the system presents a linear behavior.

For quadcopters, one can follow the following steps for each direction separately to determine the desired output trajectory that is appropriate for closed-loop identification: i) consider a sinusoidal trajectory at small frequencies and apply the PD controllers given in \eqref{eq:PDController}; ii) compute Fourier transform of the control input and of the position of the quadcopter; iii) gradually increase the frequency of the sinusoidal trajectory until the corresponding component disappears in the Fourier transform of the quadcopter's position (this step would determine the bandwidth of the system); and iv) gradually increase the magnitude of the sinusoidal trajectory (while satisfying the spatial limits) until frequency components other than that of the considered trajectory appear in the Fourier transform of the control input (this step would determine the saturation levels for the control inputs so as to ensure a linear behavior).

Following the above-mentioned procedure, we select the following desired sinusoidal trajectory to identify the dynamics of the Parrot Bebop 2:
 \begin{align}\label{eq:reference}
&\text{Desired Trajectory}=\frac{1}{10}\Big( 3  \sin(0.2 \pi t) \nonumber\\
&+0.6\sin(0.4 \pi t)+ 0.1\sin(0.7 \pi t) + 0.1\sin(\pi t)\Big),
\end{align}
which has frequency components at 0.1, 0.2, 0.35, and 0.5 Hz. We consider the desired trajectory given in \eqref{eq:reference} separately for each direction, while the desired trajectory for the other two directions is set to zero. Fig. \ref{fig:FourierTransform} presents the Fourier transform of control input and of the position of the Parrot Bebop 2 along each direction. As seen in this figure, the selected desired trajectory ensures that the Parrot Bebop 2 behaves as a linear system. According to the Fourier analysis, we set the saturation levels for the control inputs $u_x$ and $u_y$ to 0.06 [rad], and for the control input $u_z$ to 0.6 [m/s] (i.e., maximum pitch and roll angles are $\pm$4 [deg], and maximum vertical velocity is $\pm$0.6 [m/s]). Note that although these numbers are smaller than the manufacturer's recommendation, they ensure a linear behavior for the Parrot Bebop 2 drone. Furthermore, we conclude from the Fourier analysis that -3B-bandwidth of the Parrot Bebop 2 in all directions is lower than 0.5 Hz.

\begin{figure}
    \begin{center}
    \includegraphics[width=\columnwidth]{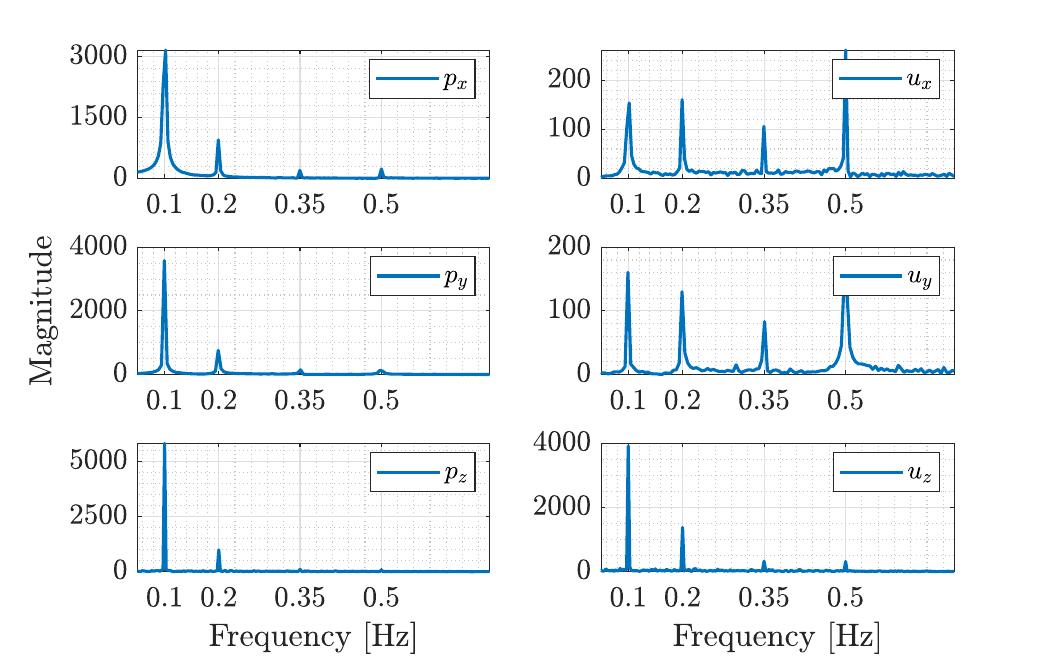}
    \caption{Fourier transform of positions and corresponding control inputs in X, Y, and Z direction.} 
    \label{fig:FourierTransform}
    \end{center}
\end{figure}

\item\textbf{Data Collection:} According to model \eqref{eq:System1}, to identify the system parameters $\alpha_i,\beta_i,~i\in\{x,y,z\}$, one needs to observe velocity and acceleration in each direction for a given control input. Since motion capture systems usually provide only position data, one would need to use numerical derivation to compute velocity and acceleration data. Note that although measurement noise in motion capture systems is small, numerical derivation can amplify the impact of the noise on velocities and accelerations, and consequently can compromise the accuracy of the identified parameters. Hence, it becomes imperative to properly filter data.

In our experiments, we use the PD control law given in \eqref{eq:PDController} and the desired identification trajectory given in \eqref{eq:reference} to collect training data. Fig. \ref{fig:3} shows the obtained data in all directions, where the Newton's difference quotient is used to obtain the velocity and acceleration data. Note that in the identification process, we use the unfiltered position and velocity data. Whereas, to obtain the acceleration data, we first use a low-pass filter with a cutoff frequency of 5 Hz to filter the velocity data, then employ the Newton's difference quotient to generate acceleration data, and finally use a low-pass filter with a cutoff frequency of 10 Hz to smoothen the acceleration data.

\begin{figure}
\begin{center}
\centering
\includegraphics[width=\columnwidth]{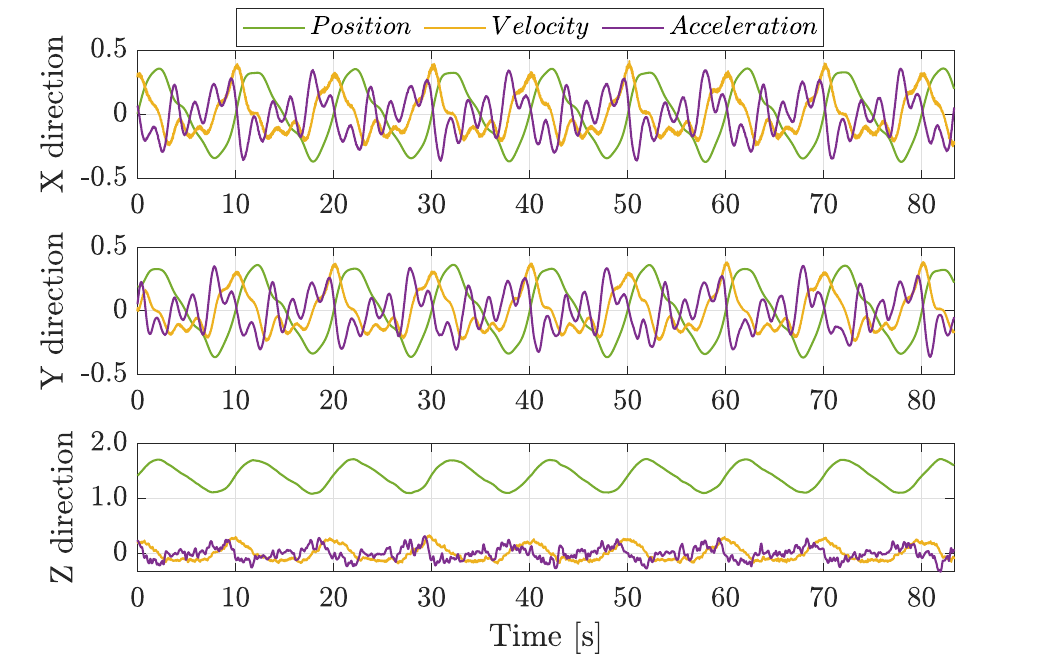}\\
\includegraphics[width=\columnwidth]{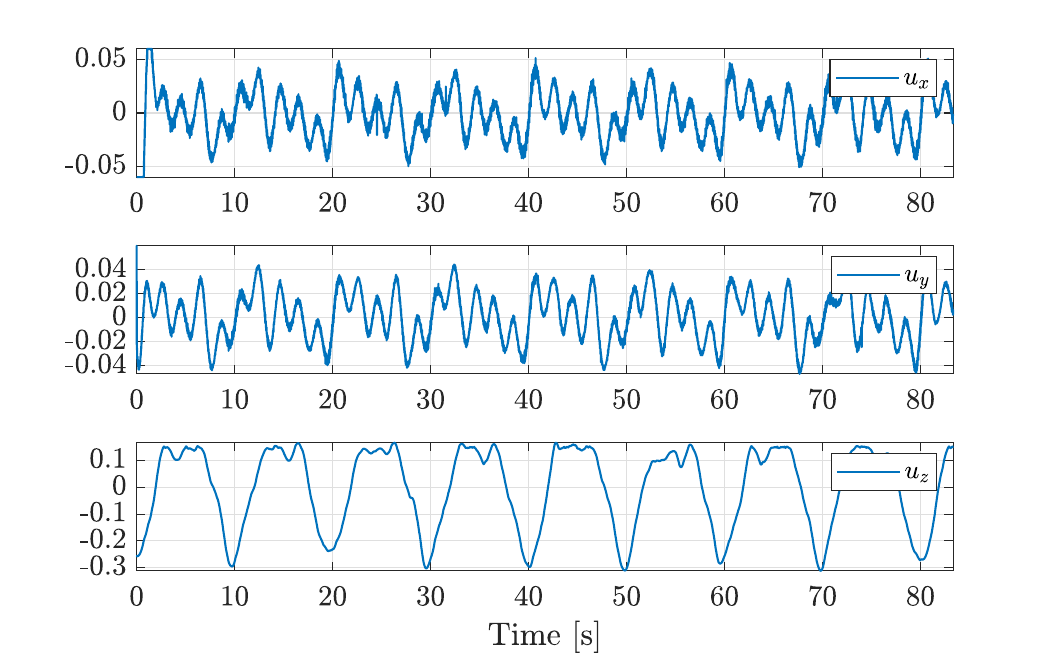}
\caption{Obtained data in X, Y, and Z directions and control input along each direction}.
\label{fig:3}
\end{center}
\end{figure}

\item\textbf{Parameter Identification:} Once position, velocity, and acceleration data is collected, model \eqref{eq:System1} suggests that one can use the linear regression techniques (see, e.g., \citep{Kleinbaum2013}) to identify the system parameters $\alpha_i,\beta_i,~i\in\{x,y,z\}$.

\begin{table}[!t]
\caption{Identified parameters for the Parrot Bebop 2 drone.}
\centering
\begin{tabular}{c|c}
Parameter &         Identified Value   \\
\hline\hline
$\alpha_x$ & 0.0527 \\
\hline
$\alpha_y$ & 0.0187 \\
\hline
$\alpha_z$ & 1.7873 \\
\hline
$\beta_x$ & -5.4779 \\
\hline
$\beta_y$ & -7.0608 \\
\hline
$\beta_z$ & -1.7382  
\end{tabular}
\label{tab:SystemParamteres}
\end{table}


We use the least square method to identify the parameters of the Parrot Bebop 2 drone. The obtained values are reported in Table \ref{tab:SystemParamteres}. Note that the obtained parameters satisfy our practical observations. For instance, let $p_x(0)=0$, $u_x(t)=0$, and $\dot{p}_x(0)=\sigma$ for some $\sigma\in\mathbb{R}_{>0}$; that is, the Parrot Bebop 2 has an initial velocity along X direction, while the roll angle is zero. According to model \eqref{eq:System1}, $p_x(t)$ is given by:
\begin{align}
p_x(t)=\frac{\sigma}{\alpha_x}\left(1-e^{-\alpha_xt}\right),~t\geq0,
\end{align}
which implies that the Parrot Bebop 2 moves toward the point $\frac{\sigma}{\alpha_x}$ in X direction, while its velocity $\dot{p}_x(t)$ converges to zero (note that $\alpha_x\in\mathbb{R}_{>0}$); this behavior is consistent with our practical observations. Also, setting $p_x(0)=\dot{p}_x(0)=0$ and $u_x(t)=\sigma$ for some $\sigma\in\mathbb{R}_{>0}$ yields:
\begin{align}
p_x(t)=\frac{\sigma}{\alpha_x}t+\frac{\sigma}{\alpha_x^2}\left(e^{-\alpha_xt}-1\right),~t\geq0,
\end{align}
which indicates that when the roll angle is kept at $\sigma$ radian, the Parrot Bebop 2 will stay in motion (with a constant steady-state velocity) in X direction; this behavior is also consistent with our practical observations.



The obtained parameters satisfy our frequency-domain observations as well. Fig. \ref{fig:Bode} compares the magnitude of the obtained transfer function at frequencies 0.1, 0.2, 0.35, and 0.5 Hz with the one computed from the Fourier transform of practical data shown in Fig. \ref{fig:FourierTransform}. As seen in Fig. \ref{fig:Bode}, the identified model approximately fits the data in all X, Y, and Z directions. 

\end{itemize}

\section{Experimental Validation: MPC-Based Navigation}\label{sec:Control}
Suggested by prior work, e.g., \citep{Huang2000,Huang2003}, this section aims at experimentally validating the identified model by investigating the performance of a MPC control scheme developed based upon the identified model. For this purpose, this section develops a steady-state-aware MPC \citep{SSMPC} for system \eqref{eq:StateEqaution} with parameters given in Subsection \ref{sec:IdentificationProcedure}. Our motivation to use the steady-state-aware MPC is that it is capable of addressing output tracking, steady-state convergence requirements, and constraint satisfaction, despite the limitations of computing power.

\begin{figure}
\begin{center}
\includegraphics[width=\columnwidth]{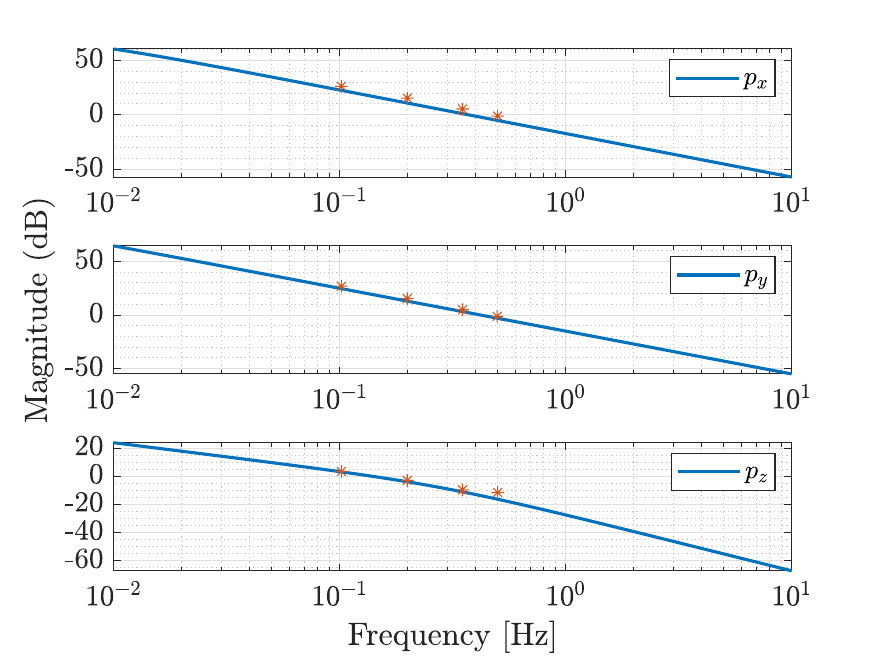}\\ 
\caption{Bode plot of the identified model for each direction (blue lines) and the magnitudes computed based on Fourier transforms (red asterisks).} 
\label{fig:Bode}
\end{center}
\end{figure}



\subsection{System Discretization}
The first step to design a steady-state-aware MPC is to discretize system \eqref{eq:StateEqaution} with an appropriate sampling period, to obtain the following discrete-time linear system:
\begin{subequations}\label{eq:DiscreteMatrices}
\begin{align}
x(k+1)=&A_dx(k)+B_du(k),\\
y(k)=&C_dx(k),
\end{align}
\end{subequations}
where the interval $[k,k+1)$ is equal to the considered sampling period.

For the Parrot Bebop 2 drone, we use a sampling period of 0.2 seconds. Note that the bandwidth of the system is 0.5 Hz (see Subsection \ref{sec:IdentificationProcedure}), and thus, the considered sampling frequency satisfies the Nyquist criterion \citep{Ogata1995}. The obtained $A_d$, $B_d$, and $C_d$ matrices are:
\begin{subequations}
\begin{align}
&A_d=\begin{bmatrix}
1 & 0.19895 & 0 & 0 & 0 & 0 \\
0 & 0.98952  & 0 & 0 & 0 & 0 \\
0 & 0 & 1.000 & 0.19963 & 0 & 0 \\
0 & 0 & 0 & 0.99627 & 0 & 0 \\
0 & 0 & 0 & 0 & 1.000 & 0.16816\\
0 & 0 & 0 & 0 & 0 & 0.69946
\end{bmatrix},\\
&B_d=\begin{bmatrix}
-0.10917348 & 0 & 0  \\
-1.08982035 & 0 & 0  \\
0 & -0.141040918& 0  \\
0 & -1.409531141 & 0  \\
0 & 0 & -0.030967224  \\
0 & 0 & -0.292295416 
\end{bmatrix},\\
&C_d=\begin{bmatrix}
1 & 0 & 0 & 0 & 0 & 0 \\
0 & 0 & 1 & 0 & 0 & 0 \\
0 & 0 & 0 & 0 & 1 & 0
\end{bmatrix}.
\end{align}
\end{subequations}

\subsection{Control Structure}

Given $r=[r_1~r_2~r_3]^\top\in\mathbb{R}^3$ as the reference position of the quadcopter, since $(A_d,B_d)$ is controllable\footnote{Since $(A_d,B_d)$ is controllable, for any given $r$, there exists \citep{Muske1993} a steady state $\mathbf{x}_f$ and steady input $\mathbf{u}_f$ such that $\mathbf{x}_f=A_d\mathbf{x}_f+B_d\mathbf{u}_f$ and $r=C_d\mathbf{x}_f$.}, it is always possible \citep{SSMPC,Limon2008} to characterize the reference signal $r$, corresponding steady state $\mathbf{x}_f$, and steady input $\mathbf{u}_f$ as $r=L\theta$, $\mathbf{x}_f=M\theta$ and $\mathbf{u}_f=W\theta$, where $L\in\mathbb{R}^{3\times3}$, $M\in\mathbb{R}^{6\times3}$, $W\in\mathbb{R}^{3\times3}$, and $\theta\in\mathbb{R}^3$ is the characterizing vector.

Given the prediction horizon $N\in\mathbb{Z}_{>0}$, the steady-state-aware MPC computes the optimal characterizing vector $\theta^\ast(k)$ and the optimal control sequence $\mathbf{u}^\ast(k):=\left[\begin{matrix}\left(u^\ast(0 | k\right)^{\top}& \cdots & \left(u^\ast(N-1 | k)\right)^{\top}\end{matrix}\right]^{\top} \in \mathbb{R}^{3N}$ at any time instant $k$ by solving the following optimization problem:
\begin{subequations}\label{eq:OptimizationProblemMain}
\begin{align}
&\theta^\ast(k),\mathbf{u}^\ast(k)=\arg\,\min _{\mathbf{u},\theta} \Big( \sum_{s=0}^{N-1}\left\Vert\hat{x}(s|k)-M\theta\right\Vert_{Q_x}^2\nonumber\\
&+\sum_{s=0}^{N-1}\left\Vert u(s|k)-W\theta\right\Vert_{Q_u}^2+\left\Vert\hat{x}(N|k)-M\theta\right\Vert_{Q_N}^2\nonumber\\
&+\left\Vert r-L\theta\right\Vert_{Q_r}^2+\left\Vert M\theta-x_{des}\right\Vert_{Q_{fx}}^2+\left\Vert W\theta-u_{des}\right\Vert_{Q_{fu}}^2\Big),\label{eq:costfunction}
\end{align}
subject to the following constraints\footnote{Note that since the purpose of this section is to validate the identified model, we do not consider any constraints on the states, like spatial constraints. It is noteworthy that the steady-state-aware MPC is able to handle both state and input constraints.}:
\begin{align}
& \hat{x}(s+1|k)=A_d\hat{x}(s|k)+B_du(s|k),~\hat{x}(0| k)=x(k),\\
& u(s|k) \in \mathcal{U},~s\in\{0, \cdots, N-1\},\\
& (\hat{x}(N | k), \theta) \in \Omega\label{eq:ConstraintTerminal},
\end{align}
\end{subequations}
where $Q_x=Q_x^{\top} \succeq 0$ ($Q_x\in\mathbb{R}^{6\times6}$), $Q_u=Q_u^{\top} \succ 0$ ($Q_u\in\mathbb{R}^{3\times3}$), $Q_{fx}=Q_{fx}^{\top} \succeq 0$ ($Q_{fx}\in\mathbb{R}^{6\times6}$), $Q_{fu}=Q_{fu}^{\top} \succeq 0$ ($Q_{fu}\in\mathbb{R}^{3\times3}$), $x_{des}\in\mathbb{R}^6$ and $u_{des}\in\mathbb{R}^3$ are desired steady state and steady input determined by the designer, $Q_N\succeq 0$ ($Q_N\in\mathbb{R}^{6\times6}$), $Q_r=Q_r^{\top} \succ 0$ ($Q_r\in\mathbb{R}^{3\times3}$), $\mathcal{U}$ is the constraint set on the control inputs (i.e., $\mathcal{U}=\{u||u_x|\leq0.06,|u_y|\leq0.06,|u_z|\leq0.6\}$), and $\Omega$ is the terminal constraint set (see subsection \ref{sec:Omega}).

For quadcopters with dynamical model described in \eqref{eq:StateEqaution}, the desired steady state and steady input are $x_{des}=[r_1~0~r_2~0~r_3~0]^\top$ and $u_{des}=[0~0~0]^\top$, respectively. Also, matrices $M$, $L$, and $W$ are computed as:
\begin{align}\label{eq:MAtricesMWL}
& M=\begin{bmatrix}1 & 0&0&0&0&0\\
0&0&1&0&0&0\\
0&0&0&0&1&0
\end{bmatrix}^\top,~L=\begin{bmatrix}
1&	0&	0\\
0&	1&	0\\
0&	0&	1 
\end{bmatrix},~W=\begin{bmatrix}
0&	0&	0\\
0&	0&	0\\
0&	0&	0 
\end{bmatrix}.
\end{align}

Note that since $W$ is the zero matrix and $u_{des}$ is set to zero, for quadcopters modeled as in \eqref{eq:StateEqaution}, the last term (i.e., $\left\Vert \mathbf{u}_f-u_{des}\right\Vert_{Q_{fu}}^2$) in the cost function \eqref{eq:costfunction} is zero.

\subsection{Determination of $Q_N$}\label{sec:QN}
Drawing inspiration from conventional MPC  such as those discussed in \citep{nicotra2018embedding,Hosseinzadeh2023RobustTermination}, it is advantageous to set $Q_N$ in \eqref{eq:costfunction} to the solution to the algebraic Riccati equation $Q_N=A_d^\top Q_NA_d-(A_d^\top Q_NB_d)(Q_u+B_d^\top Q_NB_d)^{-1}(B_d^\top Q_NA_d)+Q_x$.

\subsection{Determining the Terminal Constraint Set $\Omega$}\label{sec:Omega}
Let the terminal control law be $\kappa(x(k),\theta)=K(x(k)-M\theta)$, where the gain $K$ is $K=-(Q_u+B_d^\top Q_NB_d)^{-1}(B_d^\top Q_NA_d)$, where $Q_N$ is as in Subsection \ref{sec:QN}. It is easy to show that $A_d+B_dK$ is Schur. 

Following the arguments similar to \citep{SSMPC}, the terminal constraint set $\Omega$ can be computed as follows:
\begin{align}\label{eq:Omega}
\Omega=\{(x,\theta)|\hat{u}(\omega|x,\theta)\in\mathcal{U},\omega=0,\cdots,\omega^\ast\},
\end{align}
where $\omega^\ast\in\mathbb{Z}_{>0}$ can be obtained by solving a sequence of mathematical programming problems detailed in \citep{Gilbert1991}, and
\begin{align}
\hat{u}(\omega&|x,\theta)=K(A_d+B_dK)^\omega \hat{x}(N|k)\nonumber\\
&-K\sum_{j=1}^\omega(A_d+B_dK)^{j-1}B_dKM\theta-B_dKM\theta.
\end{align}

Note that the terminal constraint set $\Omega$ given in \eqref{eq:Omega} is convex and positively invariant, and can be easily computed at any time instant $k$.

\subsection{Theoretical Properties}

This subsection proves the properties of the steady-state-aware MPC designed to navigate the quadcopter. First, we show that the proposed steady-state-aware MPC is recursively feasible.

\begin{theorem}[Recursive Feasibility]\label{theorem:feasibility}
Consider system \eqref{eq:DiscreteMatrices}, which is subject to the above-mentioned constraints on the control inputs. Suppose that \eqref{eq:OptimizationProblemMain} is feasible at $k=0$. Then, it remains feasible for all $k>0$.
\end{theorem}

\begin{pf}
Suppose that MPC \eqref{eq:OptimizationProblemMain} is feasible at $k$, where the optimal characterizing vector and the optimal control sequence are denoted by $\theta^\ast(k)$ and $\mathbf{u}^\ast(k)$, respectively. Also, $(\hat{x}(N|k),\theta^\ast(k)) \in \Omega$, where $\hat{x}(N|k)$ is the terminal state given the optimal control sequence $\mathbf{u}^\ast(k)$.

Since the terminal constraint set $\Omega$ is positively invariant, it is concluded that $\kappa\big(\hat{x}(N|k),\theta^\ast(k)\big)=K(A_d+B_dK)\hat{x}(N|k)-KB_dKM\theta^\ast(k)-B_dKM\theta^\ast(k)\in\mathcal{U}$, and $\left(\hat{x}(N+1|k),\theta^\ast(k)\right)\in\Omega$. Thus, the characterizing vector $\theta^\ast(k)$ and the control sequence $\big[\left(u^\ast(1|k)\right)^{\top}~\cdots$ $\left(u^\ast(N-1|k)\right)^{\top}~ \left(\kappa\big(\hat{x}(N|k), \theta^\ast(k)\big)\right)^\top\big]^{\top}$ construct a feasible solution for the proposed steady-state-aware MPC at time instant $k+1$. Therefore, feasibility at time instant $k$ implies feasibility at time instant $k+1$. This implies that the steady-state-aware MPC given in \eqref{eq:OptimizationProblemMain} is recursively feasible, which completes the proof.  \hfill$\qed$
\end{pf}

Next, we show that the proposed steady-state-aware MPC ensures closed-loop stability. 

\begin{theorem}[Closed-Loop Stability]\label{theorem:stability}
Suppose that the steady-state-aware MPC given in \eqref{eq:OptimizationProblemMain} is used to control system \eqref{eq:DiscreteMatrices}. Then, $x(k)\rightarrow x_{des}$, $u(k)\rightarrow u_{dex}$, and $y(k)\rightarrow r$ as $k\rightarrow\infty$, where $x_{des}=[r_1~0~r_2~0~r_3~0]^\top$ and $u_{des}=[0~0~0]^\top$.
\end{theorem}

\begin{pf}
Let $J\big(\theta,\mathbf{u}|x(k)\big)$ and $J\big(\theta,\mathbf{u}|x(k+1)\big)$ be the cost functions of the steady-state-aware MPC given in \eqref{eq:OptimizationProblemMain} at time instants $k$ and $k+1$, respectively. Also, let  $\big(\theta^\ast(k),\mathbf{u}^\ast(k)\big)$ and  $\big(\theta^\ast(k+1),\mathbf{u}^\ast(k+1)\big)$ be the optimal solutions at time instants $k$ and $k+1$, respectively. According to the optimality of the solution $\big(\theta^\ast(k+1),\mathbf{u}^\ast(k+1)\big)\in\mathbb{R}^{3}\times\mathbb{R}^{3}$ at time instant $k+1$, we have:
\begin{align}\label{eq:Optimality1}
&J\big(\theta^\ast(k+1),\mathbf{u}^\ast(k+1)|x(k+1)\big)\nonumber\\
&~~~~~~~~\leq J\big(\theta^\ast(k),\mathbf{u}^\ast(k+1)|x(k+1)\big).
\end{align}

By subtracting $J\big(\theta^\ast(k),\mathbf{u}^\ast(k)|x(k)\big)$ from both sides of \eqref{eq:Optimality1}, one can get the following inequality:
\begin{align}\label{eq:Inequality1}
&J\big(\theta^\ast(k+1),\mathbf{u}^\ast(k+1)|x(k+1)\big)-J\big(\theta^\ast(k),\mathbf{u}^\ast(k)|x(k)\big)\nonumber\\
&~~~\leq J\big(\theta^\ast(k),\mathbf{u}^\ast(k+1)|x(k+1)\big)-J\big(\theta^\ast(k),\mathbf{u}^\ast(k)|x(k)\big).
\end{align}

According to \eqref{eq:costfunction}, and since $u^\ast(s+1|k)=u^\ast(s|k+1)$ and $\hat{x}(s+1|k)=\hat{x}(s|k+1),~s=0,\cdots,N-2$, we obtain that:
\begin{align}\label{eq:Inequality2}
&J\big(\theta^\ast(k+1),\mathbf{u}^\ast(k+1)|x(k+1)\big)-J\big(\theta^\ast(k),\mathbf{u}^\ast(k)|x(k)\big)\nonumber\\
&\leq \left\Vert\hat{x}(N | k+1)-M\theta^\ast(k)\right\Vert_{Q_N}^2-\left\Vert\hat{x}(N | k)-M\theta^\ast(k)\right\Vert_{Q_N}^2\nonumber\\
&+\left\Vert\hat{x}(N-1 | k+1)-M\theta^\ast(k)\right\Vert_{Q_x}^2+\left\Vert u(N-1 | k+1)\right\Vert_{Q_u}^2\nonumber\\
&-\left\Vert \hat{x}(0|k)-M\theta^\ast(k)\right\Vert_{Q_x}^2-\left\Vert u^\ast(0|k)\right\Vert_{Q_u}^2.
\end{align}

As shown in \citep{mayne2000Stability,SSMPC}, when the matrix $Q_N$ is determined as in Subsection \ref{sec:QN}, we have $\big\Vert\hat{x}(N | k+1)-M\theta^\ast(k)\big\Vert_{Q_N}^2-\big\Vert\hat{x}(N | k)-M\theta^\ast(k)\big\Vert_{Q_N}^2+\big\Vert\hat{x}(N-1 | k+1)-M\theta^\ast(k)\big\Vert_{Q_x}^2+\big\Vert u(N-1 | k+1)\big\Vert_{Q_u}^2\leq0$. Thus, it follows from \eqref{eq:Inequality2} that:
\begin{align}\label{eq:Inequality3}
&J\big(\theta^\ast(k+1),\mathbf{u}^\ast(k+1)|x(k+1)\big)-J\big(\theta^\ast(k),\mathbf{u}^\ast(k)|x(k)\big)\nonumber\\
&\leq-\big\Vert \hat{x}(0|k)-M\theta^\ast(k)\big\Vert_{Q_x}^2-\big\Vert u^\ast(0|k)\big\Vert_{Q_u}^2\leq0.
\end{align}

According to \eqref{eq:Inequality3}, it is obvious that imposing $J\big(\theta^\ast(k+1),\mathbf{u}^\ast(k+1)|x(k+1)\big)-J\big(\theta^\ast(k),\mathbf{u}^\ast(k)|x(k)\big)\equiv0$ implies that $u(k)=u_{des}=\mathbf{0}$. In the following, we show that imposing $J\big(\theta^\ast(k+1),\mathbf{u}^\ast(k+1)|x(k+1)\big)-J\big(\theta^\ast(k),\mathbf{u}^\ast(k)|x(k)\big)\equiv0$ implies that $x(k)=x_{des}$ as well.

On the one hand, as shown in \citep[Appendix II]{SSMPC}, imposing $J\big(\theta^\ast(k+1),\mathbf{u}^\ast(k+1)|x(k+1)\big)-J\big(\theta^\ast(k),\mathbf{u}^\ast(k)|x(k)\big)=0$ for $k\geq k^\dag$ (for some $k^\dag\in\mathbb{Z}_{\geq0}$) implies that the optimal cost function $J\big(\theta^\ast(k),\mathbf{u}^\ast(k)|x(k)\big)$ takes the following form for $k\geq k^\dag$:
\begin{align}\label{eq:CostEquilibrium1}
J\big(\theta^\ast(k),\mathbf{u}^\ast(k)|x(k)\big)=&\left\Vert r-L\theta^\ast(k)\right\Vert_{Q_r}^2\nonumber\\
&+\left\Vert M\theta^\ast(k)-x_{des}\right\Vert_{Q_{fx}}^2.
\end{align}

On the other hand, from optimality of the solution $\big(\theta^\ast(k),\mathbf{u}^\ast(k)\big)$ for $k\geq k^\dag$, we have:
\begin{align}\label{eq:CostEquilibrium2}
&J\big(\theta^\ast(k),\mathbf{u}^\ast(k)|x(k)\big)\leq J\big(r,\mathbf{u}^\ast(k)|x(k)\big),~k\geq k^\dag,
\end{align}

From \eqref{eq:CostEquilibrium1} and \eqref{eq:CostEquilibrium2}, we obtain that:
\begin{align}
J\big(\theta^\ast(k),\mathbf{u}^\ast(k)|x(k)\big)\leq&\left\Vert r-Lr\right\Vert_{Q_r}^2+\left\Vert Mr-x_{des}\right\Vert_{Q_{fx}}^2.
\end{align}
which according to matrices $M$ and $L$ given in \eqref{eq:MAtricesMWL}, we have:
\begin{align}\label{eq:CostEquilibrium3}
J\big(\theta^\ast(k),\mathbf{u}^\ast(k)|x(k)\big)\leq&0.
\end{align}

Since $J\big(\theta^\ast(k),\mathbf{u}^\ast(k)|x(k)\big)$ cannot be negative, it follows from \eqref{eq:CostEquilibrium3} that $\left\Vert r-L\theta^\ast(k)\right\Vert_{Q_r}^2=\left\Vert M\theta^\ast(k)-x_{des}\right\Vert_{Q_{fx}}^2=0$ for $k\geq k^\dag$, or equivalently $L\theta^\ast(k)=r$ and $M\theta^\ast(k)=x_{des}$ for $k\geq k^\dag$. This means that the system is at steady-state for $k\geq k^\dag$, and the steady output and state are $y(k)=L\theta^\ast(k)=r$ and $x(k)=M\theta^\ast(k)=x_{des}$, respectively.

According to the above-mentioned discussion, we conclude that $x(k)\rightarrow x_{des}$, $u(k)\rightarrow u_{des}$, and $y(k)\rightarrow r$ as $k\rightarrow\infty$, which completes the proof. \hfill$\qed$
\end{pf}

\subsection{Experimental Setting}
To conduct our experimental analysis, we set $N=10$, $Q_x=\text{diag}\{5,5,5,5,5\}$, $Q_u=\text{diag}\{35,20,1\}$, $Q_r=\text{diag}\{500,500,500\}$, and $Q_{fx}=\text{diag}\{1,1,1,1,1,1\}$. We implement the steady-state-aware MPC with one sample delay \citep{Roy2021,Hosseinzadeh2022_ROTEC}; that is, the control signal computed based on the measurements at sampling instant $k$ is applied to the Parrot Bebop 2 at sampling instant $k+1$. We use the \texttt{YALMIP} toolbox \citep{Lofberg2004} to perform the computations of the developed steady-state-aware MPC.



\subsection{Experimental Results}

\begin{figure}
\begin{center}
\includegraphics[width=\columnwidth]{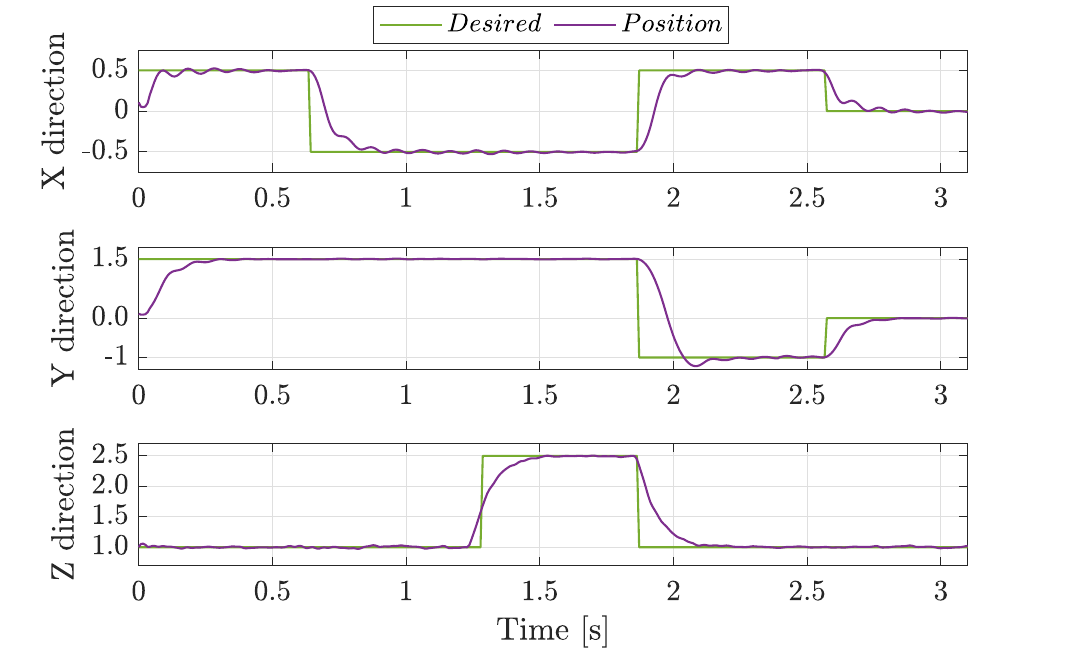}    
\caption{Set points traveled by the quadrotor Bebop 2 in the navigation experiment.} 
\label{fig:6}
\end{center}
\end{figure}

\noindent\textbf{Constant Piecewise Reference Tracking:} We assume that desired positions $p_x^d(t)$, $p_u^d(t)$, and $p_z^d(t)$ are piecewise constant. Experimental results are presented in Fig. \ref{fig:6}. As seen in this figure, the developed steady-state-aware MPC effectively steers the Parrot Bebop 2 to the desired location even in the presence of abrupt changes in the reference signal. The obtained results can serve to validate the identified parameters model for Parrot Bebop 2 which is reported in Subsection \ref{sec:IdentificationProcedure}.

\begin{figure}
\begin{center}
\includegraphics[width=8cm]{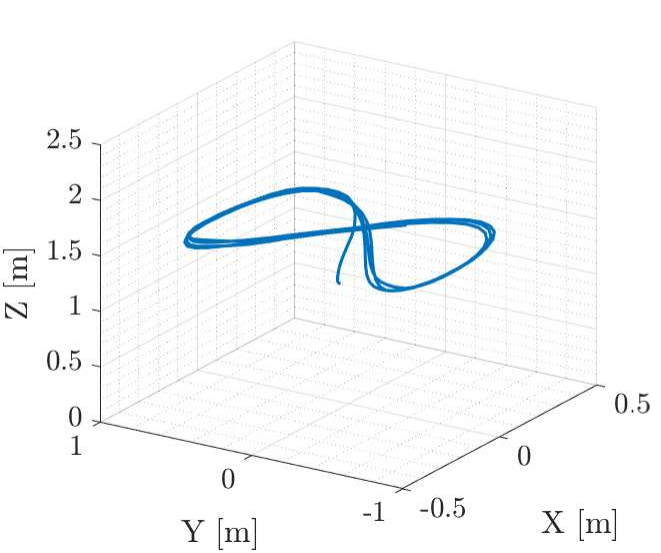}    
\caption{Path followed by the quadrotor when tracking the lemniscate of Bernoulli trajectory.} 
\label{fig:7}
\end{center}
\end{figure}

\noindent\textbf{Trajectory Tracking:} Let $p_z^d(t)=1.5$, and $p_x^d(t)$ and $P_y^d(t)$ be such that they construct the lemniscate of Bernoulli curve. Fig. \ref{fig:7} shows the trajectory traveled by the Parrot Bebop 2 in 3D. As one can see, the task of trajectory tracking with the developed steady-state-aware MPC is accomplished, implying that the identified model for the Parrot Bebop 2 is reliable.

\noindent\textbf{Discussion:} Experimental results shown in Fig. \ref{fig:6} and \ref{fig:7} affirm that although the model \eqref{eq:StateEqaution} is a simplified approximation of the dynamics of the Parrot Bebop 2, it can be safely used to design a model-based control scheme for it. Also, the computing time of the developed steady-state-aware MPC is 0.0518$\pm$0.0055 seconds, which indicates that the steady-state-aware MPC can be employed to address control objectives while ensuring constraint satisfaction even on low-end hardware.

\section{Conclusion}\label{sec:Conclusion}
This paper proposed to systematically address the complexities of quadcopters' dynamics and computational limitations. First, a physically informed modeling approach was advocated, leading to the identification of a simple yet efficient linear model for enabling onboard computation efficiency and ensuring safe navigation. Second, a steady-state-aware MPC framework was developed to effectively navigate quadcopters, while guaranteeing constraint satisfaction at all times within the confines of limited computational resources. Using a Parrot Bebop 2 as a running example, proposed methods and algorithms were evaluated and validated experimentally.

\bibliography{ref}        

\begin{thebibliography}{54}
\providecommand{\natexlab}[1]{#1}
\providecommand{\url}[1]{\texttt{#1}}
\providecommand{\urlprefix}{URL }
\expandafter\ifx\csname urlstyle\endcsname\relax
  \providecommand{\doi}[1]{doi:\discretionary{}{}{}#1}\else
  \providecommand{\doi}{doi:\discretionary{}{}{}\begingroup \urlstyle{rm}\Url}\fi

\bibitem[{Alessio and Bemporad(2009)}]{alessio2009survey}
Alessio, A. and Bemporad, A. (2009).
\newblock A survey on explicit model predictive control.
\newblock In L.~Magni, D.M. Raimondo, and F.~Allgower (eds.), \emph{Nonlinear Model Predictive Control: Towards New Challenging Applications}, 345--369. Springer Berlin, Heidelberg.

\bibitem[{Amiri and Hosseinzadeh(2024)}]{SSMPC}
Amiri, M. and Hosseinzadeh, M. (2024).
\newblock Steady-state-aware model predictive control for tracking in systems with limited computing capacity.
\newblock \emph{IEEE Control Syst. Lett.}
\newblock {DOI:}10.1109/LCSYS.2024.3370266.

\bibitem[{Bouabdallah et~al.(2004)Bouabdallah, Noth, and Siegwart}]{bouabdallah2004pid}
Bouabdallah, S., Noth, A., and Siegwart, R. (2004).
\newblock {PID} vs {LQ} control techniques applied to an indoor micro quadrotor.
\newblock In \emph{Proc. IEEE/RSJ Int. Conf. Intelligent Robots and Systems}, volume~3, 2451--2456. Sendai, Japan.

\bibitem[{Camacho and Bordons(2007)}]{camacho2007model}
Camacho, E.F. and Bordons, C. (2007).
\newblock \emph{Model predictive controllers}.
\newblock Springer.

\bibitem[{den Hof(1998)}]{Hof1998}
den Hof, P.V. (1998).
\newblock Closed-loop issues in system identification.
\newblock \emph{Annu. Rev. Control.}, 22, 173--186.

\bibitem[{Dubay and Pan(2018)}]{dubay2018distributed}
Dubay, S. and Pan, Y.J. (2018).
\newblock Distributed mpc based collision avoidance approach for consensus of multiple quadcopters.
\newblock In \emph{Proc. IEEE 14th Int. Conf. Control and Automation}, 155--160.

\bibitem[{Duggal et~al.(2016)Duggal, Sukhwani, Bipin, Reddy, and Krishna}]{Duggal2016}
Duggal, V., Sukhwani, M., Bipin, K., Reddy, G.S., and Krishna, K.M. (2016).
\newblock Plantation monitoring and yield estimation using autonomous quadcopter for precision agriculture.
\newblock In \emph{Proc. IEEE Int. Conf. Robotics and Automation}, 5121--5127. Stockholm, Sweden.

\bibitem[{Feller and Ebenbauer(2018)}]{feller2018sparsity}
Feller, C. and Ebenbauer, C. (2018).
\newblock Sparsity-exploiting anytime algorithms for model predictive control: A relaxed barrier approach.
\newblock \emph{IEEE Trans. Control Syst. Technol.}, 28(2), 425--435.

\bibitem[{Ferramosca et~al.(2008)Ferramosca, Limon, Alvarado, Alamo, and Camacho}]{Ferramosca2008}
Ferramosca, A., Limon, D., Alvarado, I., Alamo, T., and Camacho, E.F. (2008).
\newblock {MPC} for tracking with optimal closed-loop performance.
\newblock In \emph{Proc. 47th IEEE Conf. Decision and Control}. Cancun, Mexico.

\bibitem[{Ferramosca et~al.(2009)Ferramosca, Limon, Alvarado, Alamo, and Camacho}]{Ferramosca2009}
Ferramosca, A., Limon, D., Alvarado, I., Alamo, T., and Camacho, E.F. (2009).
\newblock {MPC} for tracking with optimal closed-loop performance.
\newblock \emph{Automatica}, 45(8), 1975--1978.

\bibitem[{Forssell and Ljung(1999)}]{Forssell1999}
Forssell, U. and Ljung, L. (1999).
\newblock Closed-loop identification revisited.
\newblock \emph{Automatica}, 35(7), 1215--1241.

\bibitem[{Forssell and Ljung(2000)}]{Forssell2000}
Forssell, U. and Ljung, L. (2000).
\newblock A projection method for closed-loop identification.
\newblock \emph{IEEE Trans. Autom. Control}, 45(11), 2101--2106.

\bibitem[{Gandhi and Ghosal(2018)}]{Gandhi2018}
Gandhi, D.A. and Ghosal, M. (2018).
\newblock Novel low cost quadcopter for surveillance application.
\newblock In \emph{Proc. Int. Conf. Inventive Research in Computing Applications}, 412--414. Coimbatore, India.

\bibitem[{Gilbert and Tan(1991)}]{Gilbert1991}
Gilbert, E.G. and Tan, K.T. (1991).
\newblock Linear systems with state and control constraints: the theory and application of maximal output admissible sets.
\newblock \emph{IEEE Trans. Autom. Control}, 36(9), 1008--1020.

\bibitem[{Hermand et~al.(2018)Hermand, Nguyen, Hosseinzadeh, and Garone}]{HosseinzadehMED2018}
Hermand, E., Nguyen, T.W., Hosseinzadeh, M., and Garone, E. (2018).
\newblock Constrained control of {UAVs} in geofencing applications.
\newblock In \emph{Proc. 26th Mediterranean Conf. Control and Automation}, 217--222. Zadar, Croatia.

\bibitem[{Hosseinzadeh et~al.(2022)Hosseinzadeh, Sinopoli, Kolmanovsky, and Baruah}]{Hosseinzadeh2022_ROTEC}
Hosseinzadeh, M., Sinopoli, B., Kolmanovsky, I., and Baruah, S. (2022).
\newblock {ROTEC}: Robust to early termination command governor for systems with limited computing capacity.
\newblock \emph{Syst. Control Lett.}, 161.
\newblock {Art.} no. 105142.

\bibitem[{Hosseinzadeh et~al.(2023{\natexlab{a}})Hosseinzadeh, Sinopoli, Kolmanovsky, and Baruah}]{hosseinzadeh2023TAC}
Hosseinzadeh, M., Sinopoli, B., Kolmanovsky, I., and Baruah, S. (2023{\natexlab{a}}).
\newblock Robust to early termination model predictive control.
\newblock \emph{IEEE Trans. Autom. Control}.
\newblock {DOI:} 10.1109/TAC.2023.3308817.

\bibitem[{Hosseinzadeh et~al.(2023{\natexlab{b}})Hosseinzadeh, Sinopoli, Kolmanovsky, and Baruah}]{Hosseinzadeh2023RobustTermination}
Hosseinzadeh, M., Sinopoli, B., Kolmanovsky, I., and Baruah, S. (2023{\natexlab{b}}).
\newblock Robust to early termination model predictive control.
\newblock \emph{IEEE Trans. Autom. Control}.
\newblock {DOI:} 10.1109/TAC.2023.3308817.

\bibitem[{Huang et~al.(2003)Huang, Malhotra, and Tamayo}]{Huang2003}
Huang, B., Malhotra, A., and Tamayo, E.C. (2003).
\newblock Model predictive control relevant identification and validation.
\newblock \emph{Chem. Eng. Sci.}, 58(11), 2389--2401.

\bibitem[{Huang and Tamayo(2000)}]{Huang2000}
Huang, B. and Tamayo, E.C. (2000).
\newblock Model validation for industrial model predictive control systems.
\newblock \emph{Chem. Eng. Sci.}, 55(12), 2315--2327.

\bibitem[{Ichnowski et~al.(2019)Ichnowski, Prins, and Alterovitz}]{Ichnowski2019}
Ichnowski, J., Prins, J., and Alterovitz, R. (2019).
\newblock The economic case for cloud-based computation for robot motion planning.
\newblock In \emph{Proc. 18th Int. Symp. Robotics Research}, 59--65. Puerto Varas, Chile.

\bibitem[{Idrissi et~al.(2022)Idrissi, Salami, and Annaz}]{idrissi2022review}
Idrissi, M., Salami, M., and Annaz, F. (2022).
\newblock A review of quadrotor unmanned aerial vehicles: applications, architectural design and control algorithms.
\newblock \emph{J. Intell. Robot. Syst.}, 104(2), 22.

\bibitem[{Kantue and Pedro(2018)}]{kantue2018nonlinear}
Kantue, P. and Pedro, J.O. (2018).
\newblock Nonlinear identification of an unmanned quadcopter rotor dynamics using {RBF} neural networks.
\newblock In \emph{Proc. 22nd Int. Conf. System Theory, Control and Computing}, 292--298. IEEE.

\bibitem[{Khodja et~al.(2017)Khodja, Tadjine, Boucherit, and Benzaoui}]{Khodja2017}
Khodja, M.A., Tadjine, M., Boucherit, M.S., and Benzaoui, M. (2017).
\newblock Experimental dynamics identification and control of a quadcopter.
\newblock In \emph{Proc. 6th Int. Conf. Systems and Control}, 498--502. Batna, Algeria.

\bibitem[{Kleinbaum et~al.(2013)Kleinbaum, Kupper, Nizam, and Rosenberg}]{Kleinbaum2013}
Kleinbaum, D.G., Kupper, L.L., Nizam, A., and Rosenberg, E.S. (2013).
\newblock \emph{Applied Regression Analysis and Other Multivariable Methods}.
\newblock Cengage Learning, 5 edition.

\bibitem[{Kumar et~al.(2018)Kumar, Sohail, Mahesh, and Nelakuditi}]{Kumar2018}
Kumar, K.V.V.M.S., Sohail, M., Mahesh, P., and Nelakuditi, U.R. (2018).
\newblock Crowd monitoring and payload delivery drone using quadcopter based {UAV} system.
\newblock In \emph{Proc. Int. Conf. Smart Systems and Inventive Technology}. Tirunelveli, India.

\bibitem[{Kvasnica et~al.(2019)Kvasnica, Bakar{\'a}{\v{c}}, and Klau{\v{c}}o}]{kvasnica2019complexity}
Kvasnica, M., Bakar{\'a}{\v{c}}, P., and Klau{\v{c}}o, M. (2019).
\newblock Complexity reduction in explicit {MPC}: A reachability approach.
\newblock \emph{Syst. Control Lett.}, 124, 19--26.

\bibitem[{Limon et~al.(2008)Limon, Alvarado, Alamo, and Camacho}]{Limon2008}
Limon, D., Alvarado, I., Alamo, T., and Camacho, E.F. (2008).
\newblock {MPC} for tracking piecewise constant references for constrained linear systems.
\newblock \emph{Automatica}, 44(9), 2382--2387.

\bibitem[{Lofberg(2004)}]{Lofberg2004}
Lofberg, J. (2004).
\newblock {YALMIP}: a toolbox for modeling and optimization in {MATLAB}.
\newblock In \emph{Proc. IEEE Int. Conf. Robotics and Automation}, 284--289. Taipei, Taiwan.

\bibitem[{{MATLAB}(2019)}]{MATLAB}
{MATLAB} (2019).
\newblock Parrot drone support from matlab.
\newblock \url{https://www.mathworks.com/hardware-support/parrot-drone-matlab.html}.
\newblock [Accessed \today].

\bibitem[{Mayne et~al.(2000)Mayne, Rawlings, Rao, and Scokaert}]{mayne2000Stability}
Mayne, D.Q., Rawlings, J.B., Rao, C.V., and Scokaert, P.O. (2000).
\newblock Constrained model predictive control: Stability and optimality.
\newblock \emph{Automatica}, 36(6), 789--814.

\bibitem[{Muske and Rawlings(1993)}]{Muske1993}
Muske, K.R. and Rawlings, J.B. (1993).
\newblock Model predictive control with linear models.
\newblock \emph{AIChE J.}, 39(2), 262--287.

\bibitem[{Nicotra et~al.(2018)Nicotra, Liao-McPherson, and Kolmanovsky}]{nicotra2018embedding}
Nicotra, M.M., Liao-McPherson, D., and Kolmanovsky, I.V. (2018).
\newblock Embedding constrained model predictive control in a continuous-time dynamic feedback.
\newblock \emph{IEEE Trans. Autom. Control}, 64(5), 1932--1946.

\bibitem[{Ogata(1995)}]{Ogata1995}
Ogata, K. (1995).
\newblock \emph{Discrete-Time Control Systems}.
\newblock Pearson, 2 edition.

\bibitem[{Pairan et~al.(2020)Pairan, Shamsudin, and Zulkafli}]{pairan2020neural}
Pairan, M.F., Shamsudin, S.S., and Zulkafli, M.F. (2020).
\newblock Neural network based system identification for quadcopter dynamic modelling: A review.
\newblock \emph{J. Adv. Mech. Eng. Appl.}, 1(2), 20--33.

\bibitem[{Pannek and Worthmann(2011)}]{Pannek2011}
Pannek, J. and Worthmann, K. (2011).
\newblock Reducing the prediction horizon in {NMPC}: An algorithm based approach.
\newblock In \emph{Proc. IEEE Int. Conf. Robotics and Automation}, 7969--7974. Taipei, Taiwan.

\bibitem[{Pinto et~al.(2020)Pinto, Marciano, Bacheti, Moreira, Brand{\~a}o, and Sarcinelli-Filho}]{pinto2020high}
Pinto, A.O., Marciano, H.N., Bacheti, V.P., Moreira, M.S.M., Brand{\~a}o, A.S., and Sarcinelli-Filho, M. (2020).
\newblock High-level modeling and control of the bebop 2 micro aerial vehicle.
\newblock In \emph{Proc. Int. Conf. Unmanned Aircraft Systems}, 939--947. Athens, Greece.

\bibitem[{Roy et~al.(2021)Roy, Hobbs, Anderson, Caccamo, and Chakraborty}]{Roy2021}
Roy, D., Hobbs, C., Anderson, J.H., Caccamo, M., and Chakraborty, S. (2021).
\newblock Timing debugging for cyber\textemdash physical systems.
\newblock In \emph{Proc. 2021 Design, Automation and Test in Europe Conference and Exhibition}, 1893--1898. Grenoble, France.

\bibitem[{Saif and Eminoglu(2022)}]{Saif2022}
Saif, E. and Eminoglu, I. (2022).
\newblock Modelling of quad-rotor dynamics and hardware-in-the-loop simulation.
\newblock \emph{J. Eng.}, 2022(10), 937--950.

\bibitem[{Santana et~al.(2014)Santana, Brandao, Sarcinelli-Filho, and Carelli}]{Santana2014}
Santana, L.V., Brandao, A.S., Sarcinelli-Filho, M., and Carelli, R. (2014).
\newblock A trajectory tracking and {3D} positioning controller for the {AR.Drone} quadrotor.
\newblock In \emph{Proc. Int. Conf. Unmanned Aircraft Systems}, 756--767. Orlando, FL, USA.

\bibitem[{Santos et~al.(2019)Santos, Rosales, Sarapura, Sarcinelli-Filho, and Carelli}]{Santos2019}
Santos, M.C.P., Rosales, C.D., Sarapura, J.A., Sarcinelli-Filho, M., and Carelli, R. (2019).
\newblock An adaptive dynamic controller for quadrotor to perform trajectory tracking tasks.
\newblock \emph{J. Intell. Robot. Syst.}, 93, 5--16.

\bibitem[{Santos et~al.(2017{\natexlab{a}})Santos, Rosales, Sarcinelli-Filho, and Carelli}]{Santos2017_2}
Santos, M.C.P., Rosales, C.D., Sarcinelli-Filho, M., and Carelli, R. (2017{\natexlab{a}}).
\newblock A novel null-space-based {UAV} trajectory tracking controller with collision avoidance.
\newblock \emph{IEEE/ASME Trans. Mechatron.}, 22(6), 2543--2553.

\bibitem[{Santos et~al.(2017{\natexlab{b}})Santos, Santana, Brandao, Sarcinelli-Filho, and Carelli}]{Santos2017}
Santos, M.C., Santana, L.V., Brandao, A.S., Sarcinelli-Filho, M., and Carelli, R. (2017{\natexlab{b}}).
\newblock Indoor low-cost localization system for controlling aerial robots.
\newblock \emph{Control Eng. Pract.}, 61, 93--111.

\bibitem[{Sawma et~al.(2018)Sawma, Khatounian, Monmasson, Ghosn, and Idkhajine}]{sawma2018effect}
Sawma, J., Khatounian, F., Monmasson, E., Ghosn, R., and Idkhajine, L. (2018).
\newblock The effect of prediction horizons in {MPC} for first order linear systems.
\newblock In \emph{Proc. IEEE Int. Conf. Industrial Technology}, 316--321. Lyon, France.

\bibitem[{Shauqee et~al.(2021)Shauqee, Rajendran, and Suhadis}]{shauqee2021quadrotor}
Shauqee, M.N., Rajendran, P., and Suhadis, N.M. (2021).
\newblock Quadrotor controller design techniques and applications review.
\newblock \emph{INCAS Bulletin}, 13(3), 179--194.

\bibitem[{Shead et~al.(2010)Shead, Muske, and Rossiter}]{Shead2010}
Shead, L.R.E., Muske, K.R., and Rossiter, J.A. (2010).
\newblock Conditions for which linear {MPC} converges to the correct target.
\newblock \emph{J. Process Control}, 20(10), 1243--1251.

\bibitem[{Sun et~al.(2019)Sun, Dai, Liu, Dimarogonas, and Xia}]{sun2019robust}
Sun, Z., Dai, L., Liu, K., Dimarogonas, D.V., and Xia, Y. (2019).
\newblock Robust self-triggered {MPC} with adaptive prediction horizon for perturbed nonlinear systems.
\newblock \emph{IEEE Trans. Autom. Control}, 64(11), 4780--4787.

\bibitem[{Tang et~al.(2017)Tang, Xiao, and Li}]{tang2017nonlinear}
Tang, Y.R., Xiao, X., and Li, Y. (2017).
\newblock Nonlinear dynamic modeling and hybrid control design with dynamic compensator for a small-scale uav quadrotor.
\newblock \emph{Meas.}, 109, 51--64.

\bibitem[{Wang et~al.(2020)Wang, Huang, Wen, Rodriguez, Garcia, Gooi, and Zeng}]{wang2020event}
Wang, B., Huang, J., Wen, C., Rodriguez, J., Garcia, C., Gooi, H.B., and Zeng, Z. (2020).
\newblock Event-triggered model predictive control for power converters.
\newblock \emph{IEEE Trans. Ind. Electron.}, 68(1), 715--720.

\bibitem[{Wang et~al.(2023)Wang, Zhou, Duan, Zhao, Cai, Zhai, Liu, and Ren}]{wang2023nonlinear}
Wang, M., Zhou, J., Duan, X., Zhao, D., Cai, P., Zhai, J., Liu, X., and Ren, C. (2023).
\newblock Nonlinear system identification for quadrotors with neural ordinary differential equations.
\newblock In \emph{Proc. IEEE Int. Conf. Unmanned Systems}, 317--322.

\bibitem[{Wang et~al.(2016)Wang, Man, Cao, Zheng, and Zhao}]{Wang2016}
Wang, P., Man, Z., Cao, Z., Zheng, J., and Zhao, Y. (2016).
\newblock Dynamics modelling and linear control of quadcopter.
\newblock In \emph{Proc. Int. Conf. Advanced Mechatronic Systems}, 498--503. Melbourne, VIC, Australia.

\bibitem[{Wang et~al.(2017)Wang, Ramirez-Jaime, Xu, and Puig}]{wang2017nonlinear}
Wang, Y., Ramirez-Jaime, A., Xu, F., and Puig, V. (2017).
\newblock Nonlinear model predictive control with constraint satisfactions for a quadcopter.
\newblock \emph{J. Phys. Conf. Ser.}, 783(1), 012025.

\bibitem[{Yousfi and Tournier(1991)}]{Yousfi1991}
Yousfi, C. and Tournier, R. (1991).
\newblock Steady state optimization inside model predictive control.
\newblock In \emph{Proc. American Control Conf.}, 1866--1870. Boston, MA, USA.

\bibitem[{Zhao and Go(2014)}]{Zhao2014}
Zhao, W. and Go, T.H. (2014).
\newblock Quadcopter formation flight control combining {MPC} and robust feedback linearization.
\newblock \emph{J. Frank. Inst.}, 351(3), 1335--1355.

\end{thebibliography}

\end{document}